\documentclass[conference]{IEEEtran}
\IEEEoverridecommandlockouts
\usepackage{cite}
\usepackage{amsmath,amssymb,amsfonts}
\usepackage{algorithmic}
\usepackage{graphicx}
\usepackage{subfigure}
\usepackage{textcomp}
\usepackage[linesnumbered,ruled]{algorithm2e}
\usepackage{array}
\usepackage{float}

\newlength\mylen

\newcolumntype{M}[1]{>{\centering\arraybackslash}m{#1}}
  
\def\BibTeX{{\rm B\kern-.05em{\sc i\kern-.025em b}\kern-.08em
    T\kern-.1667em\lower.7ex\hbox{E}\kern-.125emX}}
\begin{document}

\title{Improving Route Choice Models by Incorporating Contextual Factors via Knowledge Distillation
{\footnotesize \textsuperscript{}}
}


\author{\IEEEauthorblockN{Qun Liu\IEEEauthorrefmark{1},
Supratik Mukhopadhyay\IEEEauthorrefmark{1}, Yimin Zhu\IEEEauthorrefmark{1},
Ravindra Gudishala\IEEEauthorrefmark{1}, \\ Sanaz Saeidi\IEEEauthorrefmark{1}, Alimire Nabijiang\IEEEauthorrefmark{1}}

\IEEEauthorblockA{\IEEEauthorrefmark{1}
Louisiana State University, Baton Rouge, LA, USA\\
\{qun, supratik\}@csc.lsu.edu, \{yiminzhu, rgudis1, ssaeid1, anabij1\}@lsu.edu}
}


\maketitle

\begin{abstract}
Route Choice Models predict the route choices of travelers traversing an urban area. Most of the route choice models link route characteristics of alternative routes to those chosen by the drivers. The models play an important role in prediction of traffic levels on different routes and thus assist in development of efficient traffic management strategies that result in minimizing traffic delay and maximizing effective utilization of transport system. High fidelity route choice models are required to predict traffic levels with higher accuracy. Existing route choice models do not take into account dynamic contextual conditions such as the occurrence of an accident, the socio-cultural and economic background of drivers, other human behaviors, the dynamic personal risk level, etc. As a result, they can only make predictions at an aggregate level and for a fixed set of contextual factors. For higher fidelity, it is highly desirable to use a model that captures significance of subjective or contextual factors in route choice. This paper presents a novel approach for developing high-fidelity route choice models with increased predictive power by augmenting existing aggregate level  baseline models with information on drivers' responses to contextual factors obtained from Stated Choice Experiments carried out in an Immersive Virtual Environment through the use of knowledge distillation. 

\end{abstract}


\section{Introduction}
It is widely known that traffic congestion has significant environmental, economic, and public health consequences. The total cost and time loss associated with traffic congestion in the US has been reported to be more than \$121 billion per year and 38 hours per person, respectively \cite{schrank2009urban}. In the US, people mostly prefer to use freeways and highways. However in case of traffic congestions, alternative routes are also taken to avoid travel delay \cite{Ravindra,shojaat2016sustained,ben2010road}. Mainstream research shows growing interest and need for better understanding drivers’ route choice behavior \cite{prato2009route, liu2017safernet, ben2003discrete,lima2016understanding,park2007learning}.

Route Choice Models \cite{ben1985discrete,di2016boundedly,pursula1993urban,khattak1993commuters,prato2009route,bovy2012route} predict the route choices of travelers traversing an urban area. Most of the route choice models link route characteristics of alternative routes to those chosen by the drivers. The models play an important role in prediction of traffic levels on different routes and thus assist in development of efficient traffic management strategies that result in minimizing traffic delay and maximizing effective utilization of transport system. 

\begin{figure}[t!]
    \centering
    \includegraphics[width=0.33\textwidth]{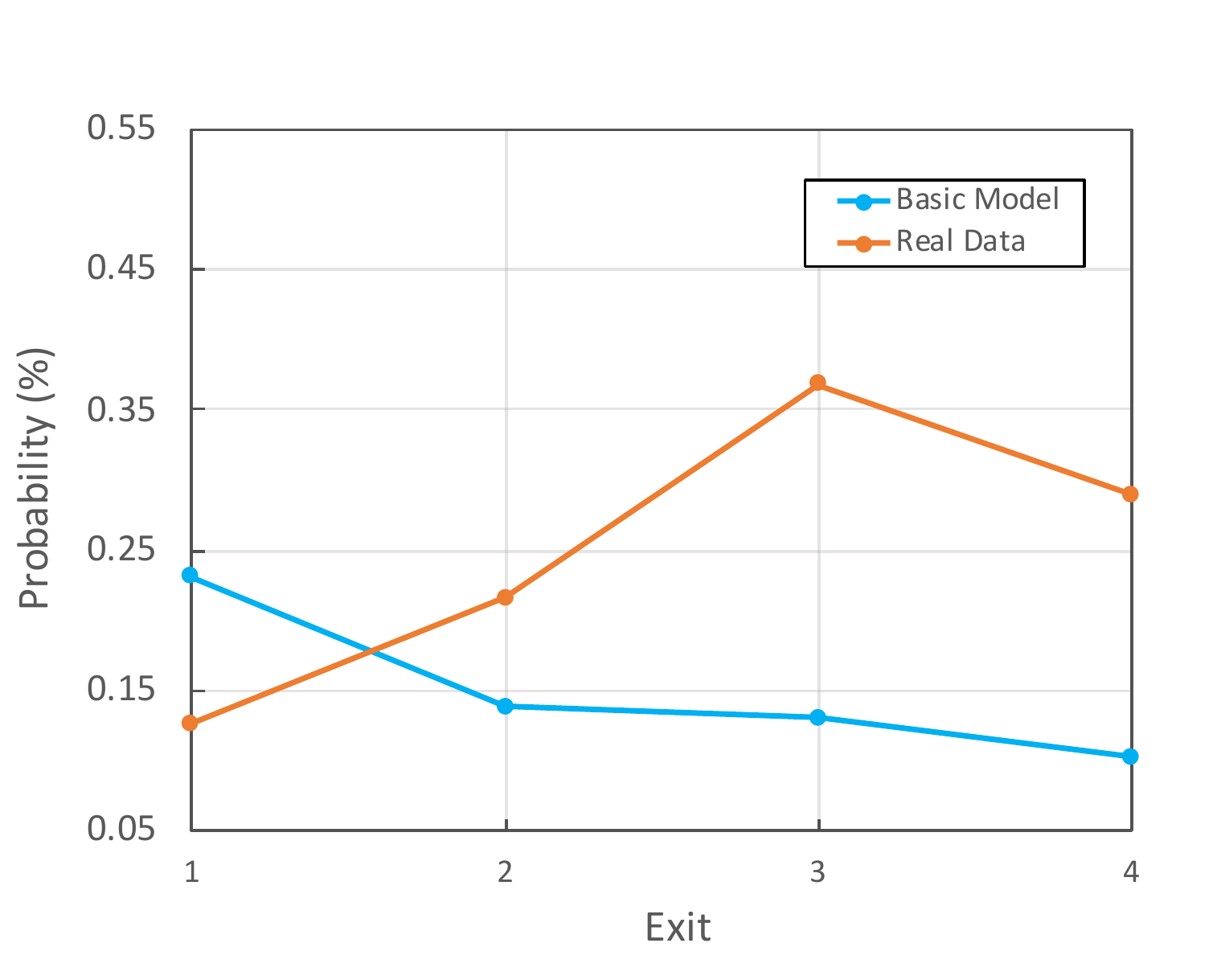}
    \caption{Comparison of Predictions Made by an Aggregate Route Choice Model with Ground Truth.}
    \label{comp_result1}
\end{figure}

High fidelity route choice models are required to predict traffic levels with higher accuracy.  Existing route choice models use revealed preference behavior to model route choice. The use of revealed choice data limits the accuracy of the prediction as it fails to capture subjective factors of drivers at individual level and allows prediction only at an aggregate level.   Fig.\ref{comp_result1} shows the route choice predictions made by a basic aggregate level route choice model (blue line) compared with real data collected from the field (red line). More precisely, Fig.\ref{comp_result1} shows the probability of drivers exiting a freeway segment through one of the four  available exits as predicted by a baseline aggregate route choice model (blue line); the red line in Fig.\ref{comp_result1}  shows the ground truth. It can be seen from Fig.\ref{comp_result1} that the predictions made by the basic model deviate widely from the ground truth. Existing route choice models do not take into account dynamic contextual conditions such as the occurrence of an accident,  the socio-cultural and economic background of drivers,  other human behaviors,  the dynamic personal risk level,  etc. As a result, they can only make predictions at an aggregate level and for a fixed set of contextual factors. Therefore, for higher fidelity, it is highly desirable to use a  methodology that captures significance of subjective or contextual factors in route choice. 

\begin{figure*}[t!]
\centering
\includegraphics[width=0.98\textwidth]{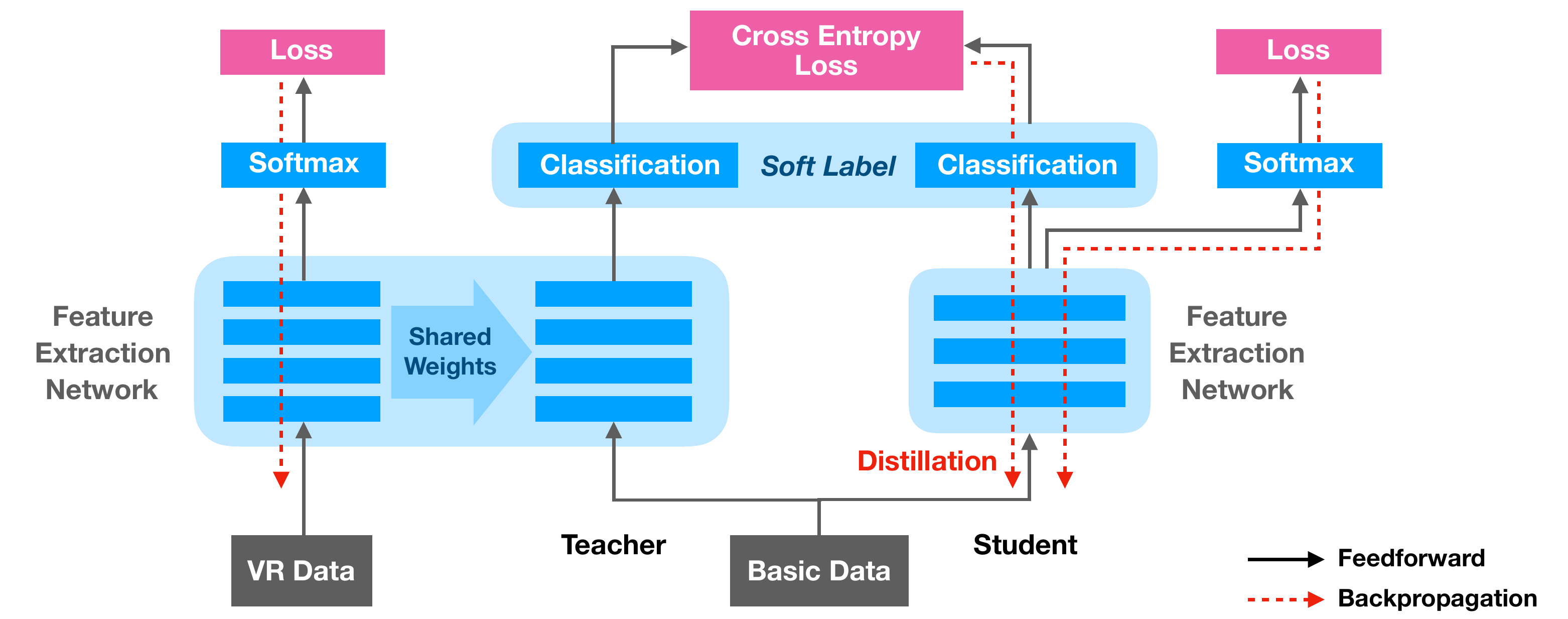}
\caption{Overview of our framework architecture.}
\label{fig:bf}
\end{figure*} 

Adding subjective or contextual  requires availability of the data at individual or disaggregate level. Stated Choice Experiments (SCEs) are a scientific  methodology  to capture the effect of context sensitive factors in route choice. The current advancements in virtual reality technology can enhance stated choice experiments by allowing  researchers to present them in a realistic manner that enhances the realism of the experiments and allows one to elicit information about route choice  made by a driver. Integrated Virtual Environments (IVEs) \cite{weidner2017comparing,ihemedu2017virtual} provide a good platform to conduct SCE and elicit responses to route choice experiments as realistically as possible. The promise of IVE applications in collecting data includes, but  is not limited to, providing a safe and user-friendly experimental platform, being inexpensive and highly portable, as well as generating context-aware and high-fidelity data.

This paper presents a novel approach for developing high-fidelity route choice models with increased predictive power by augmenting existing aggregate level  baseline models with information on drivers' responses to contextual factors  obtained from SCE carried out in an IVE through the use of knowledge distillation. Our approach uses the prior knowledge acquired by a teacher neural network pretrained on data about drivers' responses to contextual factors to augment a student neural network (a baseline model) in a guided fashion. We  demonstrate experimentally that the predictions of  the augmented model are much closer to reality than that of  the baseline. 

\paragraph*{Contributions} The paper makes the following contributions.
\begin{itemize}
    \item It presents a novel approach  using knowledge distillation for developing high-fidelity route choice models by augmenting  existing baseline models with information about drivers' reaction to contextual factors acquired from SCEs in IVEs. 
    \item We present a general end-to-end knowledge distillation
framework that uses a multilayer perceptron  as a  feature extraction network
to provide a feature learning architecture for teacher and
student networks and  then transfers  knowledge from the former to the latter by
optimizing  distillation loss.
\end{itemize}

\section{Related Work}\label{rw}
In this section, we discuss related work on route choice models and knowledge distillation frameworks. 
\paragraph{Route Choice Models} Transportation engineers have been studying commuter route choice behavior for four decades now. Engineers  developing route choice models theorized that travel time plays a crucial and important role in  the selection of a route.  Route choice behavior theories began to evolve in  the late eighties and early nineties as engineers' understanding of route choice behavior improved by studying data about  empirical route choice behavior. Pursula and Talvite \cite{pursula1993urban} developed a mathematical route model by postulating that drivers do consider other factors apart from travel time in making a route choice. In \cite{khattak1993commuters,bonsall1990drivers}, the authors  discovered that commuters prefer to use habitual routes when traveling in familiar areas as opposed to choosing a route that provides them with maximum utility. Other researchers such as Doherty and Miller \cite{doherty2000computerized} investigating route choice found that apart from travel time, factors such as residential location, familiarity with the route, and employment locations are significant in the route choice process. 

\paragraph{Knowledge Distillation} Deep learning techniques have achieved success in a variety of tasks \cite{krizhevsky2012imagenet,NeuralNetwork,transactions,character,deeplearningsurvey,DeepSAT}. Authors in \cite{bucilua2006model, liu2018unsupervised}  showed that it is possible to compress  heavy models of ensembles, that require large amount of  storage and computational power, to a single small and efficient model without significant loss. Hinton et. al. \cite{hinton2015distilling} proposed a different compression technique for knowledge distillation in a  neural network.  This method  showed excellent performance in distilling knowledge from heavy models.

Adaptive distillation loss (ADL) was proposed in \cite{tang2019learning} for single-stage detectors in the knowledge distillation setting. It magnifies distillation loss for ``harder'' examples while reducing the same for ``easier'' ones. 



In \cite{ge2019low}, the authors proposed a method for selective knowledge distillation  for solving the problem of low-resolution face recognition in the wild with low computational and memory cost. In \cite{yim2017gift}, the authors showed that knowledge distillation can help optimize neural networks. In \cite{mutual}, the authors present a method for training smaller neural networks by distilling knowledge from a  relatively bigger network. In \cite{question}, the authors present a framework for training models with specialized information  in the question-answering space through  knowledge distillation. 

\begin{figure}[t!]
    \centering
    \includegraphics[width=0.35\textwidth]{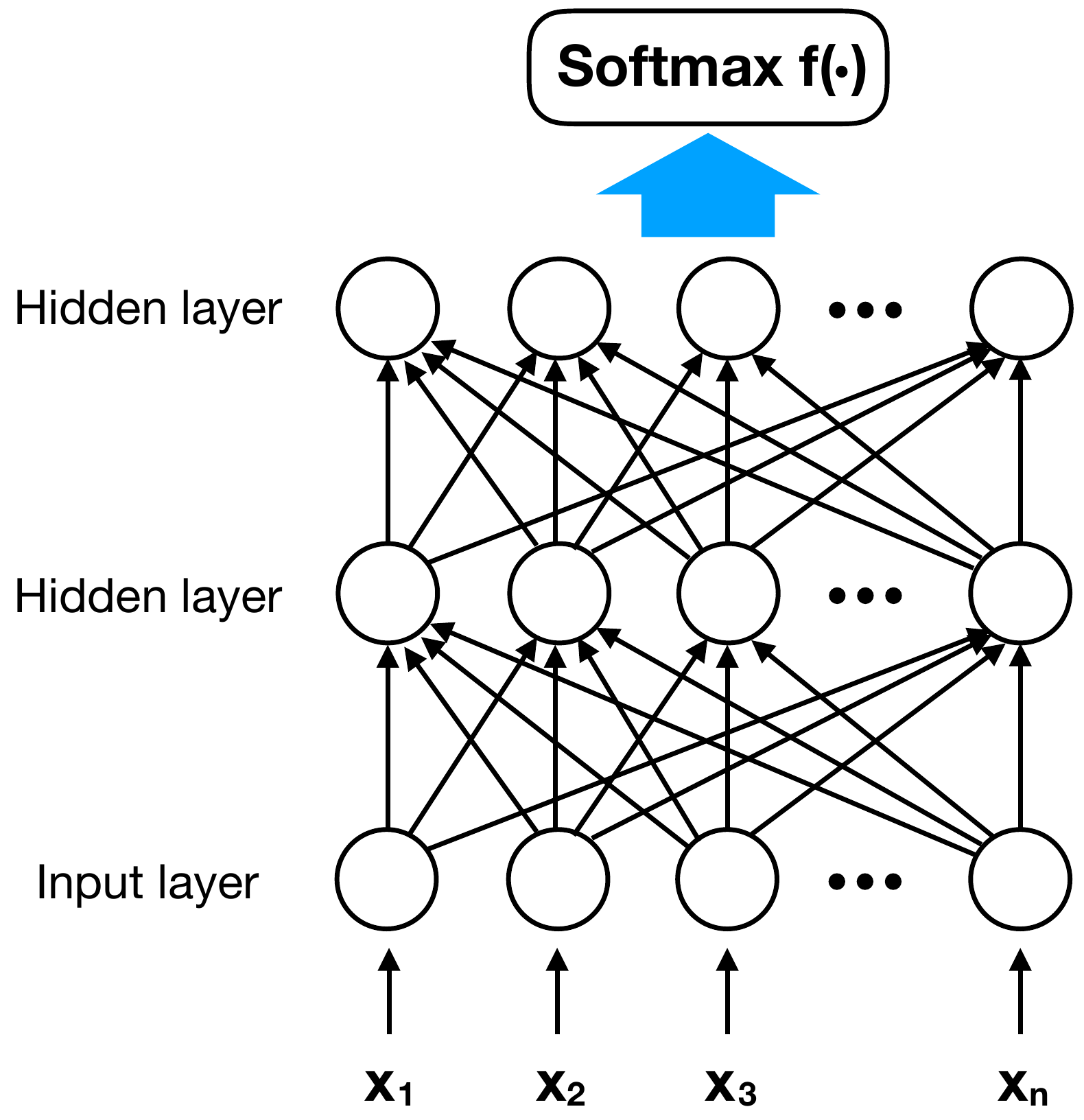}
    \caption{The illustration for an example multilayer perceptron structure of a feature extraction network. The output from the feature extraction network then fed into softmax layer for computing probabilities. }
    \label{fig:mlp}
\end{figure}

\section{Proposed Method}\label{pm}
In this section, we present an overview of our proposed method. This is followed by a description of the IVE  for SCE. Finally, we present our knowledge distillation-based framework for developing high-fidelity route choice models in Sections \ref{ann} and \ref{kd}.

\subsection{Overview}\label{ov}
We present a framework for developing high-fidelity route choice models with increased predictive power by augmenting existing aggregate level   baseline models with contextual information obtained from SCE carried out in an IVE through the use of knowledge distillation. 

The overall architecture of our framework is shown in Fig.\ref{fig:bf}. As shown in the figure, for both the teacher and the student we use feature extraction networks (see below).   The teacher is first pre-trained on data acquired  from SCE in IVE (called \emph{VR data}; see below). The \emph{basic data} (see below) acquired from predictions by the baseline route choice  model is partitioned into a training set and a test set. The training set is   used for training a student as well as for distilling knowledge  to the student from a teacher through knowledge distillation. The student is evaluated on the test set. 

\textbf{Basic  Route Choice Model and Data} We  considered a basic  mathematical route choice model  adapted from one of the common route choice models \cite{ben2004route} in literature  that serves as the baseline model. 


The baseline model predicts the probability ($P_b$) of exiting  a highway through a given exit using the following equation,
\begin{equation}\label{bm}
P_b = \alpha_b T
\end{equation}
where  the constant $\alpha_b$ is 0.601 \cite{ben2004route}, and $T$ is the travel time on the alternative route to a  fixed destination after exiting the highway. We  uniformly sampled  10,000 driving record samples  from the probability  distribution predicted by the baseline model for a highway segment with four exits, along with associated  travel times for the  alternative routes after exiting the highway through the available exits, for a fixed destination. We call this dataset the \emph{basic data}.

\subsection{Immersive Virtual Environment and VR Data}\label{vr}
Immersive virtual environments (IVEs) are popular for  simulating  environment spaces and objects in 3D and provide immersion in  interactive surroundings  with abundant multimedia information.  We created multiple scenarios for a  highway segment having four exits, together with embedded contextual factors  (such as traffic conditions: normal, medium, heavy) to collect  data from participant volunteer drivers about their interactions with these factors. These  interactions together with driver demographic  backgrounds, are  hard to acquire in the real world. 

The dataset acquired from the IVE environment was augmented by  using a  Gaussian mixture model to learn the overall distribution from which IID (independent and identically distributed) samples are drawn. We call this set of samples the \emph{VR Data}. 


\subsection{Multilayer Perceptron Models for Feature Extraction}\label{ann}


Let  $X \in \mathbb{R}^n$ be the input features to a  fully connected MLP (see Fig.\ref{fig:mlp}) $M(\cdot)$ with $l$ hidden layers (not including  a softmax layer; only two hidden layers shown in Fig.\ref{fig:mlp}), called the \emph{feature extraction network}.  Let  the parameters of $M(\cdot)$ be denoted by  $\theta_m$. Let the output response vector for  $M(\cdot)$ be  $y$.  Let the  neurons in the hidden layers  of   $M(\cdot)$ have the  nonlinear activation function $g(\cdot)$. Then the feature vector $F$ from output layer of  $M(\cdot)$  can be described by the following equation.
\begin{equation}
F = M(g(X); \theta_m)
\end{equation}
For the feature extraction network  $M(\cdot)$, we adopt softmax regression (as shown in Fig.\ref{fig:mlp} as  the function layer $f(\cdot)$ with  parameters $\theta_f$) to obtain  probabilities of exiting through the available  ``highway exits".  The probability that an individual driver  exits  through  highway exit $k$ ($k \in 1 \ldots,4$) for a feature vector $F$ (that characterizes the individual as well as the contextual factors they are subjected to) is given by the following equation. 
\begin{equation}
P(y=k \mid F) = f^k(F;\theta_f)
\end{equation}
For training $M(\cdot)$, we use the cross-entropy loss, described by the following equation.
\begin{equation}
\mathcal{L}_F(F,y) = -\sum_k y^k \log{P(y=k\mid F)}
\end{equation}
where $F$ is the feature vector, and $y$ is the response vector. 


\subsection{Knowledge Distillation}\label{kd}
\begin{figure}[t!]
    \centering
    \includegraphics[width=0.4\textwidth]{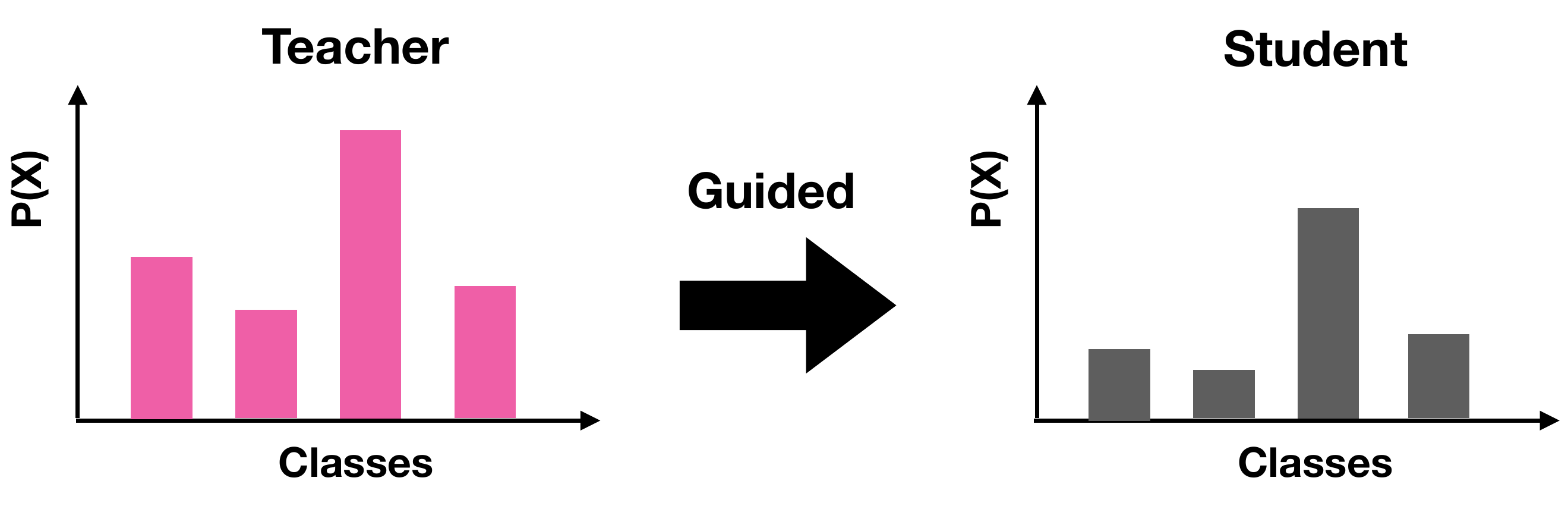}
    \caption{Distilling knowledge through the softened softmax.}
    \label{teacher_student}
\end{figure}

Knowledge distillation, also known as model compression, aims to learn a small or shallow neural network (normally called  the student model, denoted as $S(\cdot)$) with limited training examples and computational power by transferring the generalization ability from a large well-trained neural network (called  the teacher model, denoted as $T(\cdot)$), as shown in Fig.\ref{fig:bf} and Fig.\ref{teacher_student}. During training, the student model will be guided by the teacher.  The student attempts to match its softened softmax outputs with  that  of the teacher, and its hard softmax outputs with the ground truth. Given the output  $z_i$ before the last layer of a neural network, usually called logits, the softmax transforms $z_i$ to a probability $P_i$ using the following equation.
\begin{equation}
P_i = \dfrac{\exp(z_i)}{\sum_j \exp(z_j)} \label{eqp}
\end{equation}
 Normal softmax tends to set the probability for one class to one and that for the rest to zero. This makes it hard to distill  hidden knowledge to  the student with the teacher as the source.  To improve the generalization ability of student model and efficiently use the hidden knowledge,  Hinton et. al. \cite{hinton2015distilling} proposed high temperature softmax function  in lieu of using the normal softmax (temperature=1 in this case).  Then the probability $P_i$ is given by the following equation. 
\begin{equation}
P_i = \dfrac{\exp(z_i/T)}{\sum_j \exp(z_j/T)}
\end{equation}

\begin{algorithm}[b]
\caption{Training Algorithm} \label{alg1}
Pretrain($teacher$)\\
Initialize($student$) \\
\For{epoch=1,2,...,K}{
    \For{number of batches}{
        $D$ $\gets$ Sample $n$ examples with its labels $y$ \\
        $z$ $\gets$ Forward(($teacher$, $student$), $D$) \\
        $\mathcal{L}_d$ $\gets$ Loss($z$, $y$) \\
        $grad$ $\gets$ Backward($\mathcal{L}_d$) \\
        Update the parameters of student network regarding the gradients,
        $\nabla_{\theta_s} \dfrac{1}{n} \mathlarger{\sum}\limits_{i \in D} \mathcal{L}_d(i)$
    }
}
\end{algorithm}

Let the input features be denoted by $x$. Let the output from the teacher model $T(\cdot)$ with parameters $\theta_t$ be $T(x;\theta_t)$. Let  the output from the student model    $S(\cdot)$ with  parameters $\theta_s$ be $S(x;\theta_s)$. Knowledge distillation from the teacher to the student is achieved by   minimizing the distillation loss $mathcal{L}_d$, between the networks, given by the following equation,
\begin{equation}
\begin{aligned}  
\mathcal{L}_d(X,Y) = &\alpha \sum_n \mathcal{L}_1 (y_n, S(x_n;\theta_s))\\
                     & + \beta \sum_n \mathcal{L}_2 (S(x_n;\theta_s),T(x_n;\theta_t))
\end{aligned}
\end{equation}
where $X$ is a batch of training data that has $n$ training examples, $Y$ is the  vector of corresponding labels for the training data $X$. The  distillation loss $\mathcal{L}_d$ comprises of two losses, $\mathcal{L}_1$ and $\mathcal{L}_2$ (both cross entropy loss functions), which correspond to  supervised loss constrained by the ground truth data and softened loss constrained by  the teacher model, respectively. The constants $\alpha$ and $\beta$ are non-negative  and are used for balancing these two losses.  Here, we adopt the cross-entropy loss ($\mathcal{L}$) given by the following equation, 
\begin{equation}
\begin{aligned} 
\mathcal{L}(x,y) = - \dfrac{1}{N} \sum_{n=1}^{N} \sum_{m=1}^{M} [x_{nm} \log(y_{nm}) \\
+ (1-x_{nm})\log(1-y_{nm})]
\end{aligned}
\end{equation}
where $X_{nm}$ is the $m$th element of the $n$th training example, $N$ is the size of training data, and $M$ is the input size.


\subsection{Algorithm}\label{pf}
 Both teacher and student models use feature extraction networks for extracting features. The teacher and student models are MLPs with softmax layer for classification.   During each training iteration, the training set from the basic data is fed into  both the teacher model and student model. The student model is trained using backpropagation with the ground truth as the hard target and the output of the teacher as the soft target. The algorithm for knowledge distillation is shown in Algorithm \ref{alg1}. 
 

As shown in Algorithm \ref{alg1}, an end-to-end fashion training is used for our framework.  For each iteration, the algorithm computes the logits $z$ for the input data.  The cross-entropy loss is computed for backpropagating the gradients in the student network. The parameters  in  the student network are updated using gradient descent. 
\begin{equation}
\nabla_{\theta_s} \dfrac{1}{n} \mathlarger{\sum}\limits_{i \in D} \mathcal{L}_d(i)
\end{equation}

After training, the student network is tested standalone on the test set. 

\begin{table*}[htp!]
\caption{Description of the variables used in the collected virtual reality data}
\label{vrdata}
\begin{tabular}{lll}
\hline
ID     & Variable    & Description                                             \\
\hline
1      & Traffic     & 1 Normal, 2 Medium, 3 Heavy                              \\
2      & Urgency     & 1 Urgency, 2 Non-Urgency                                        \\
3      & Social impact   & 1 No, 2 Yes
\\
4      & Age     & 1 less than 25, 2 greater than or equal to 25                      \\
5      & Gender  &  1 Male, 2 Female                 
\\
6      & Race                               &  1 Middle Eastern, 2 White, 3 Other  \\
7      & Education                          &  1 Post graduate degree, 2 High school graduate, 3 College graduate                                                           \\
8      & Employment status                  &  1 Employed part time, 2 Employed full time, 3 Student, 4 Unemployed looking for work 
\\
9      & Concern while stuck in traffic     &  1 Hours of extra travel time, 2 Chaos, 3 Monetized value of delay, 4 Speed reduction due to congestion               \\
10     & Familiarity with the environment   &  1 Once a week, 2 Once a year, 3 Once a month, 4 More than once a week, 5 Never                                 \\
11     & Financial Concerns                 &  1 Sometimes, 2 Always, 3 Most of the time, 4 About half the time, 5 Never       \\
12     & Choice                             &  1 First exit, 2 Second exit, 3 Third exit, 4 Fourth exit, 5 Fifth exit     \\
\hline
\end{tabular}
\end{table*}

\section{Experimental Evaluation}\label{exp}
Experimental evaluation  will illustrate the performance of our framework with respect to predicting the probability that  an individual driver takes each of the four exits in the highway segment considered. We start with the details of data collection.
  

\subsection{Data Collection}

 The highway segment chosen for our experiment corresponds to the route of I-10 in 
 in Baton Rouge, Louisiana, United States, between  Horace Wilkinson Bridge and the intersection of Perkins Rd and Staring Ln, 
 that has four exits in the middle. 
 The travel time measured on the alternative route after taking each exit provided by {\it Google Maps} \footnote{https://www.google.com/maps/} on September 20th, 2018, are $31.7$ min, $18.9$ min, $17.8$ min, and $13.9$ min.

\begin{figure}[b!]
    \centering
    \includegraphics[width=0.42\textwidth]{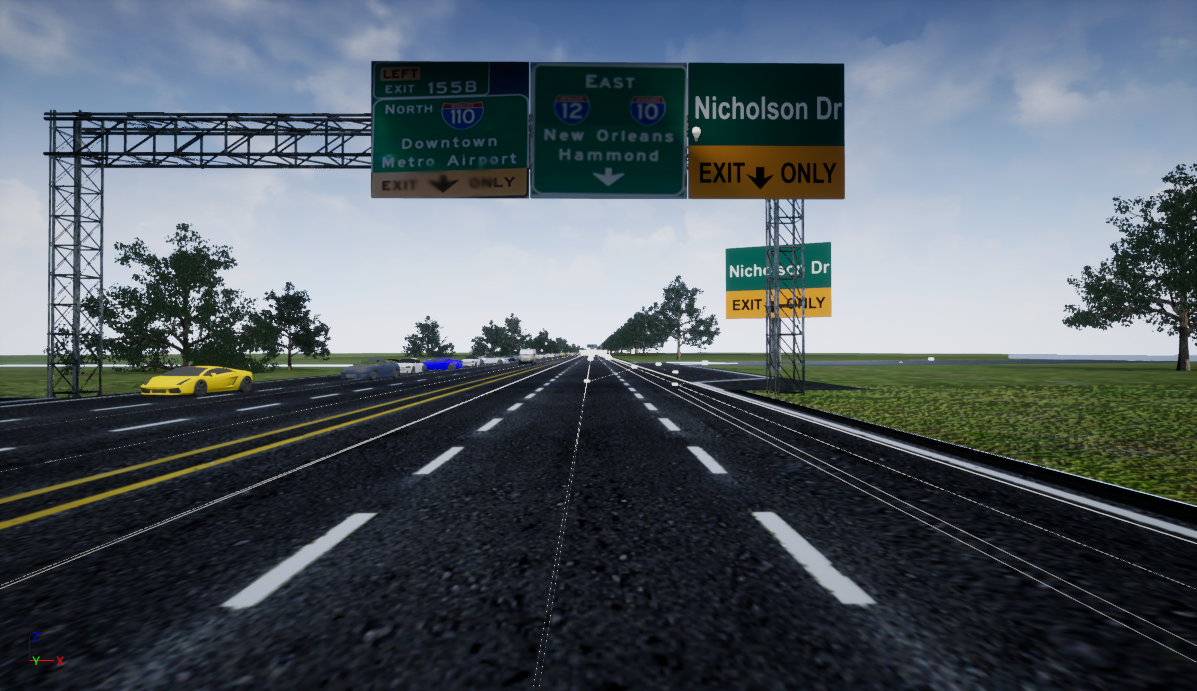}
    \caption{A Sample IVE scenario for the experiments.}
    \label{ive}
\end{figure}

\textbf{IVE Experimental Setting}
In this study, we used a driving environment (see Fig.\ref{ive}) that is  based on the I-10, starting off the Mississippi River (Horace Wilkinson)  bridge all the way to the intersection of Perkins Rd and Staring Ln. Along the way, four alternate routes were introduced to the participants, Exits 1, 2, 3, and 4, the latter of which would be College Dr.

In our Immersive Virtual Environments (IVEs), we constructed ten experimental scenarios combining sets of contextual factors (see Table \ref{vrdata} for a description of the contextual factors). 

We had forty one volunteers (20 male and 21 females; age: $31.44 \pm 7.97$)  participate in the SCE. At the beginning of the experiment, the following information was elicited from the participants through a survey: 1) demographic characteristics (age, gender, race, education, employment status); 2) top concerns while they were stuck in the traffic congestion. Their choices included hours of extra travel time, speed reduction, monetized value of delay, additional vehicle operating cost; 3) familiarity with the area; 4) socio-economic status (having concerns about spending less money on your gas).


Participants were presented with the same  origin and the destination in all the scenarios. However,  distinct  contextual factor(s) were presented in each scenario and participants were required to choose their preferred route.  

Each participant encountered each of the  driving scenarios apart from a baseline scenario. The baseline scenario was designed to  collect information about a  participant’s route choice  in a normal traffic situation with non-urgent bound condition. The contextual factors of traffic flow  were varied over three levels, i.e., normal, medium, and heavy density, in the context of a work-bound and home-bound trips; during the work-bound trip, participants were asked to consider the importance of meeting the time of arrival commitment, while no such factor was considered in the  home-bound trip. The social impact factor considered  the impact of  other drivers’ route choices on a driver's own route choice.



\begin{figure*}[t!]
\centering
\subfigure[Probabilities of leave in heavy traffic]{
    \includegraphics[width=0.3\linewidth, keepaspectratio]{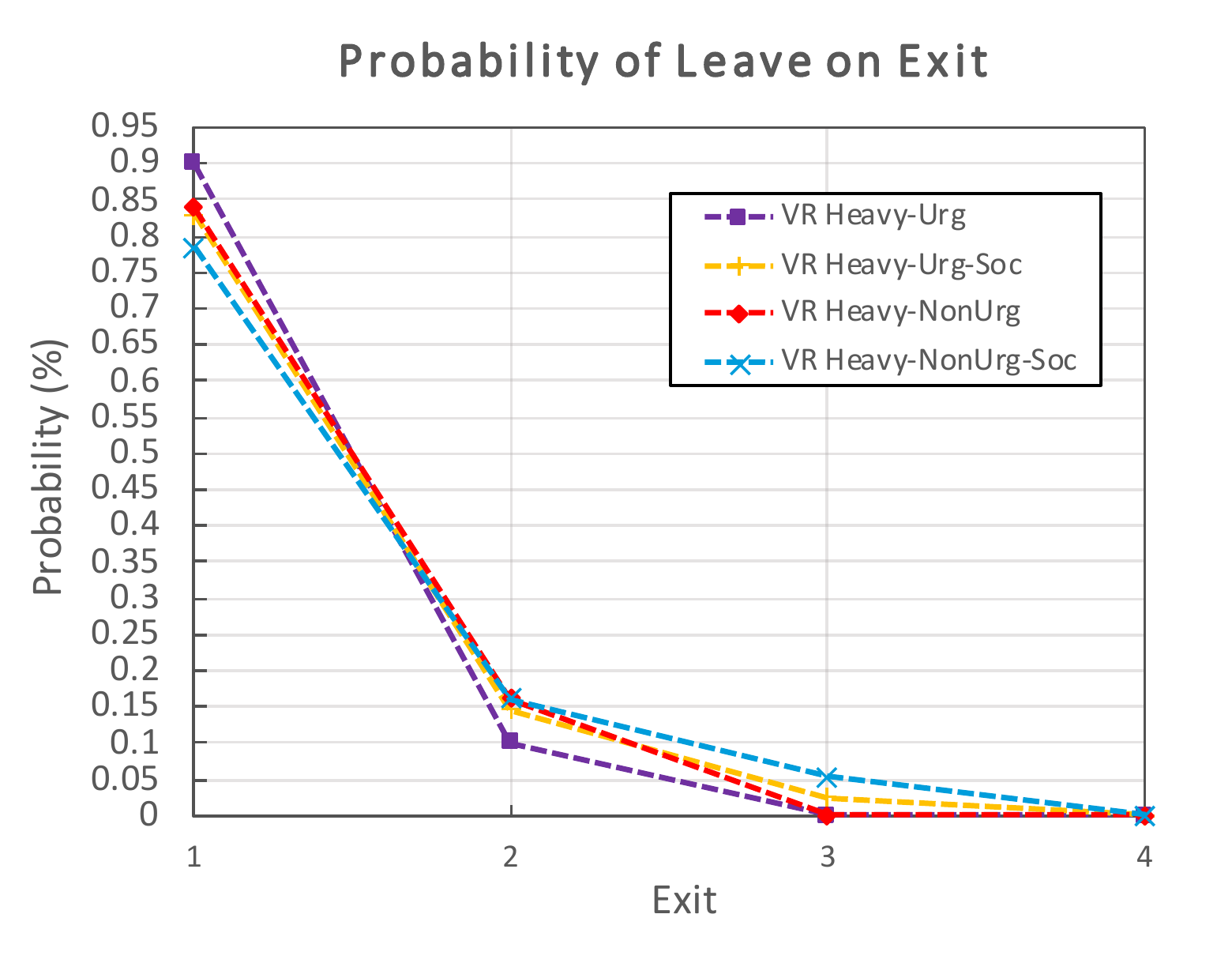}
    \label{fig:heavy}
}
\hspace{5pt}
\subfigure[Probabilities of leave in medium traffic]{
    \includegraphics[width=0.3\linewidth, keepaspectratio]{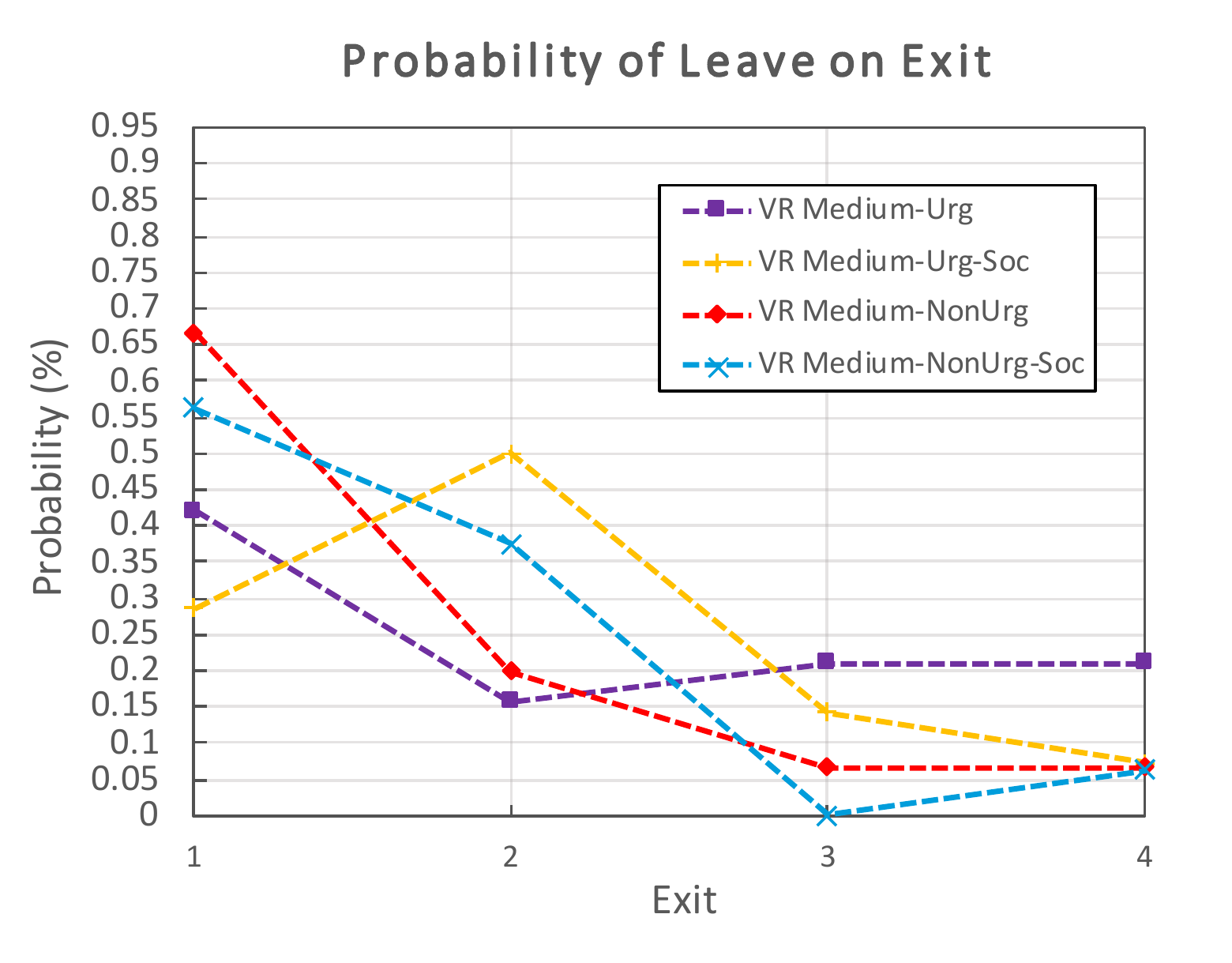}       
    \label{fig:medium}
}
\hspace{5pt}
\subfigure[Probabilities of leave in normal traffic]{
    \includegraphics[width=0.3\linewidth, keepaspectratio]{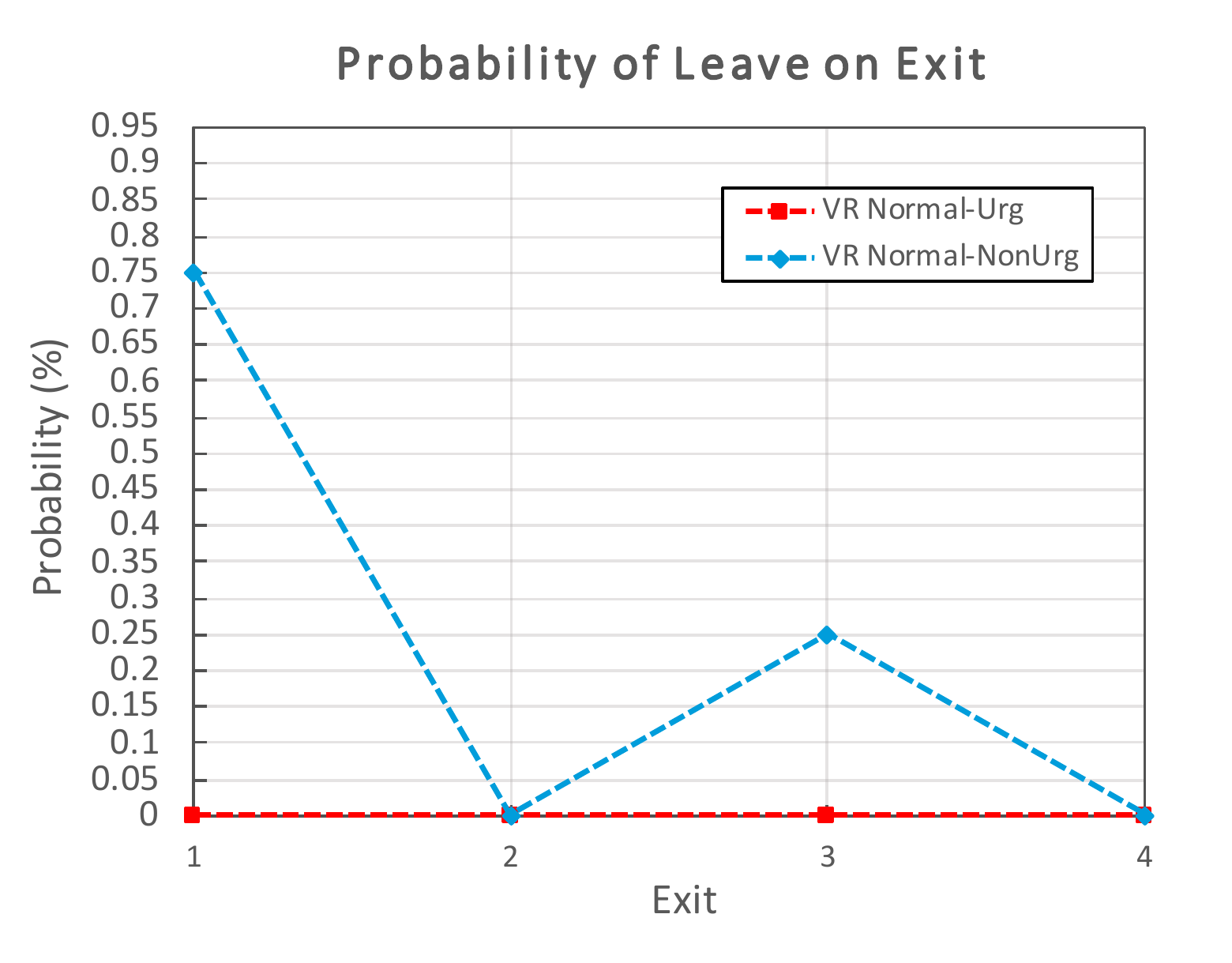}       
    \label{fig:normal}
}
\caption[Optional caption for list of figures]{The probabilities of leave calculated from the collected driving records in VR environment.}
\label{fig:vrfig}
\end{figure*}


\textbf{VR Data} From the SCE in IVE involving $41$ volunteers, a total of  410 driving records collected. Since the data collected from SCE in IVE is limited,  to better train the teacher network, we augmented  it using a Gaussian mixture model (GMM). The data collected was categorical.  We preprocessed and transformed it to the ordinal data before augmenting it using a GMM.  After data augmentation,  10,000 synthetic driving records  were generated. Each  driving record was associated with its travel time corresponding to the  alternative route for the exit taken by a driver. The 10,000 synthetic driving records together with their associated travel time is called the VR data.   The VR data was divided by $80\%$ for training and $20\%$ for testing and the training set was used to train the teacher.

For each exit, the probability of  taking it was calculated based on the data collected in the IVE as shown in Fig.\ref{fig:vrfig}. From Fig.\ref{fig:heavy}, it can be seen that the majority of drivers act consistently under heavy traffic, but use more options when they encounter medium traffic (Fig.\ref{fig:medium}).  As indicated in Fig. \ref{fig:normal}, we did not consider the social impact factor in the normal traffic scenario (that is the impact on an individual driver on seeing a large number of  drivers taking an exit) in our experiment due to the high cost in creating such a scenario. 

\textbf{Basic Data} The probabilities for the four exits are then computed according to the formula \ref{bm}. Furthermore, for the basic data, we uniformly randomly  sampled  10,000 driving records based on the probability distribution predicted by the baseline route choice model. Each record  was associated with its travel time corresponding to the  alternative route for the exit taken by a driver.  For each driving record, we assigned the  value one to the  Urgency variable  if its actual value  is less than or equal to 13 in the scale of 1 to 60. Otherwise it was assigned to two. The contextual variables, which are present  in  the VR data,  but do not occur in  the basic data are set to zero. We used this augmented dataset for knowledge distillation. We  divided this dataset into training  ($80\%$) and  testing sets ($20\%$). 

During knowledge distillation,  the teacher model, pretrained on the augmented VR data, provides the  prior  knowledge for our framework.

During training the student model on basic data, our framework incorporated the pretrained teacher model for guiding the student. Specifically, it computes the   cross-entropy loss in softened softmax function between the teacher and the student  in the backpropagation procedure  for  distilling the knowledge to student model. 
During inference, and we executed the  student model  as a standalone. It  extracts features of each test data point from dense layers then predicts the route choice probability distribution.

\begin{figure*}[t!]
\centering
\subfigure[Prediction accuracy of teacher network on VR data]{
    \includegraphics[width=0.31\linewidth, keepaspectratio]{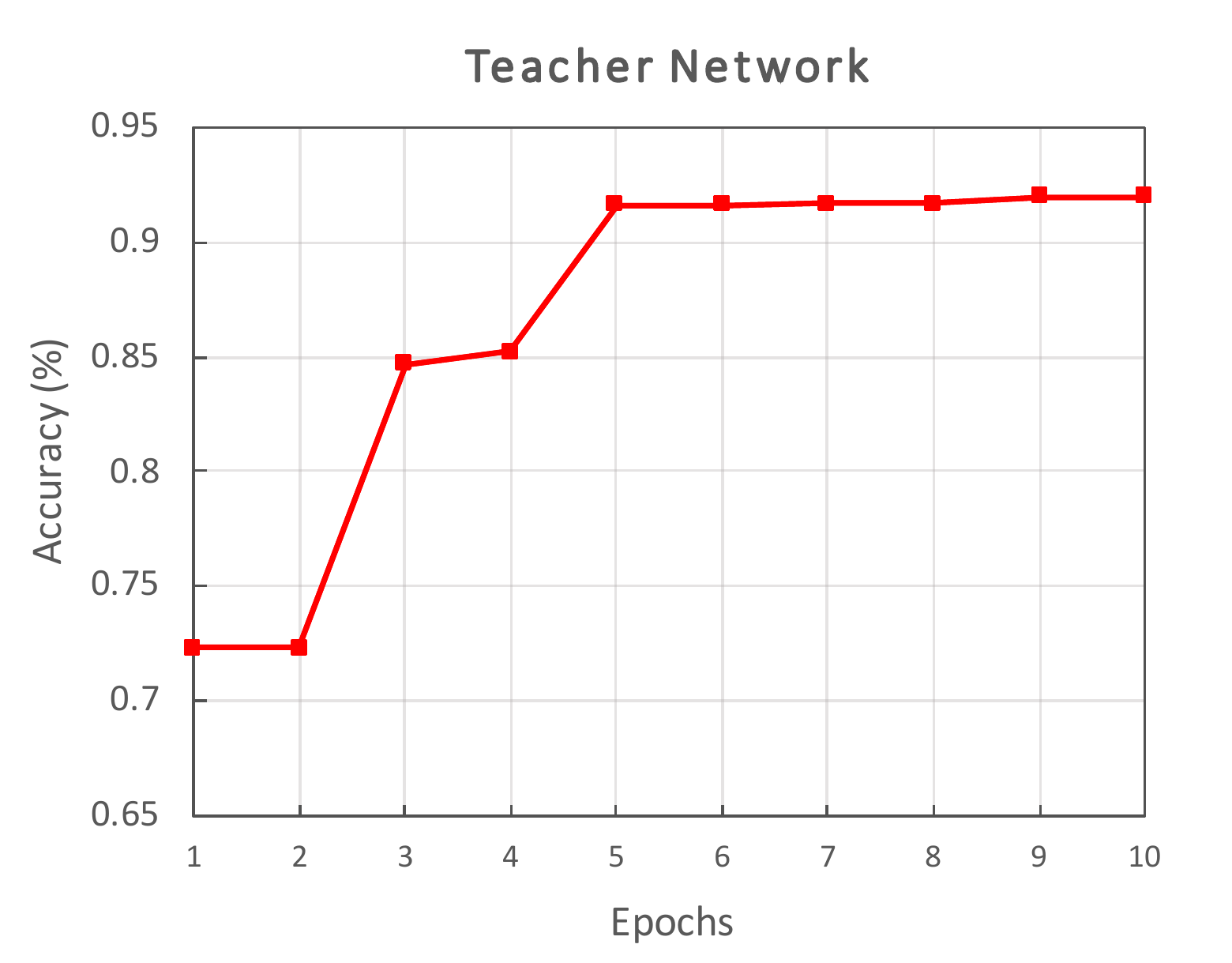}
    \label{fig:tvr}
}
\hspace{5pt}
\subfigure[Prediction accuracy of student network on Basic data]{
    \includegraphics[width=0.31\linewidth, keepaspectratio]{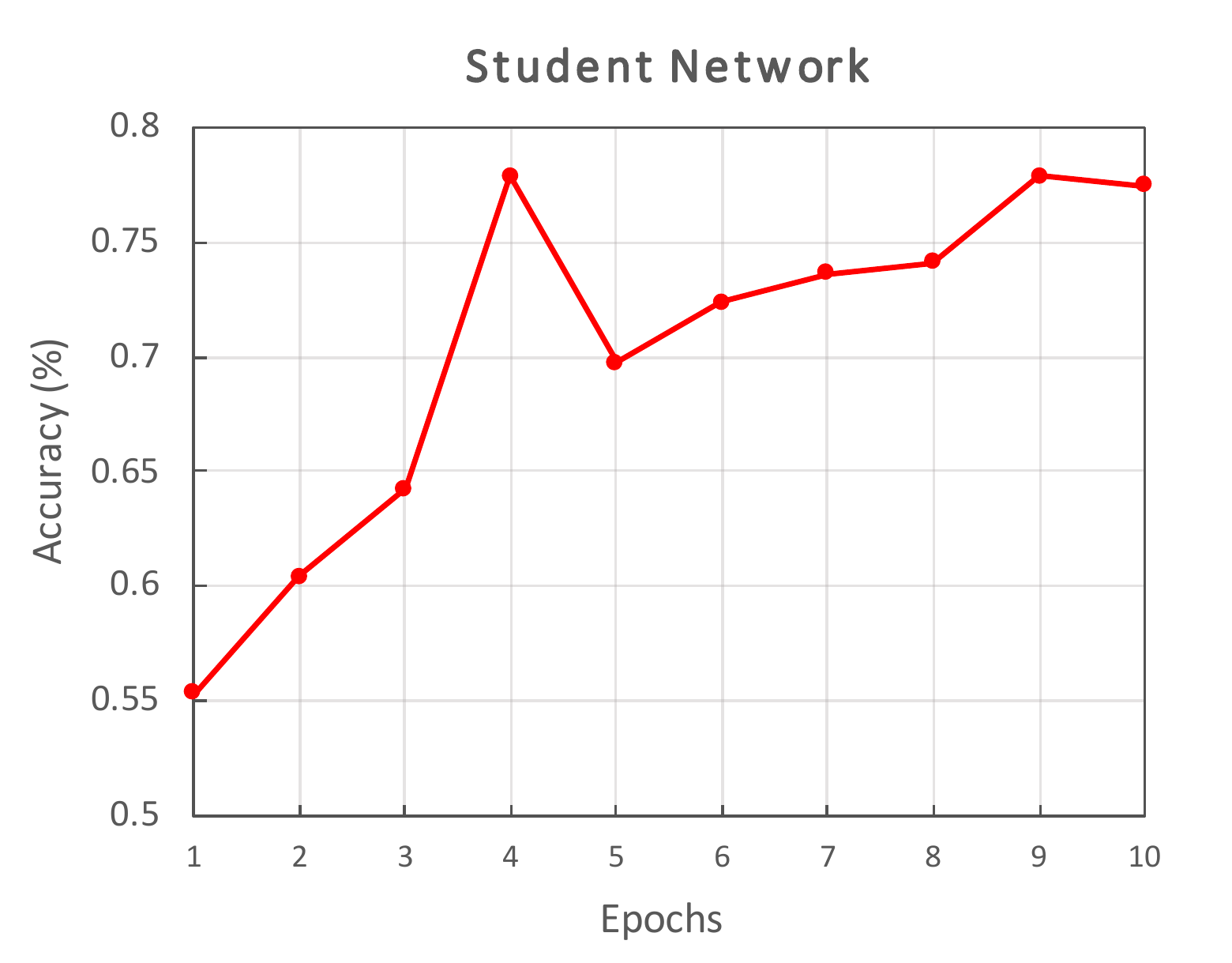}       
    \label{fig:sbasic}
}
\hspace{5pt}
\subfigure[Prediction accuracy of teacher-student network on Basic data]{
    \includegraphics[width=0.31\linewidth, keepaspectratio]{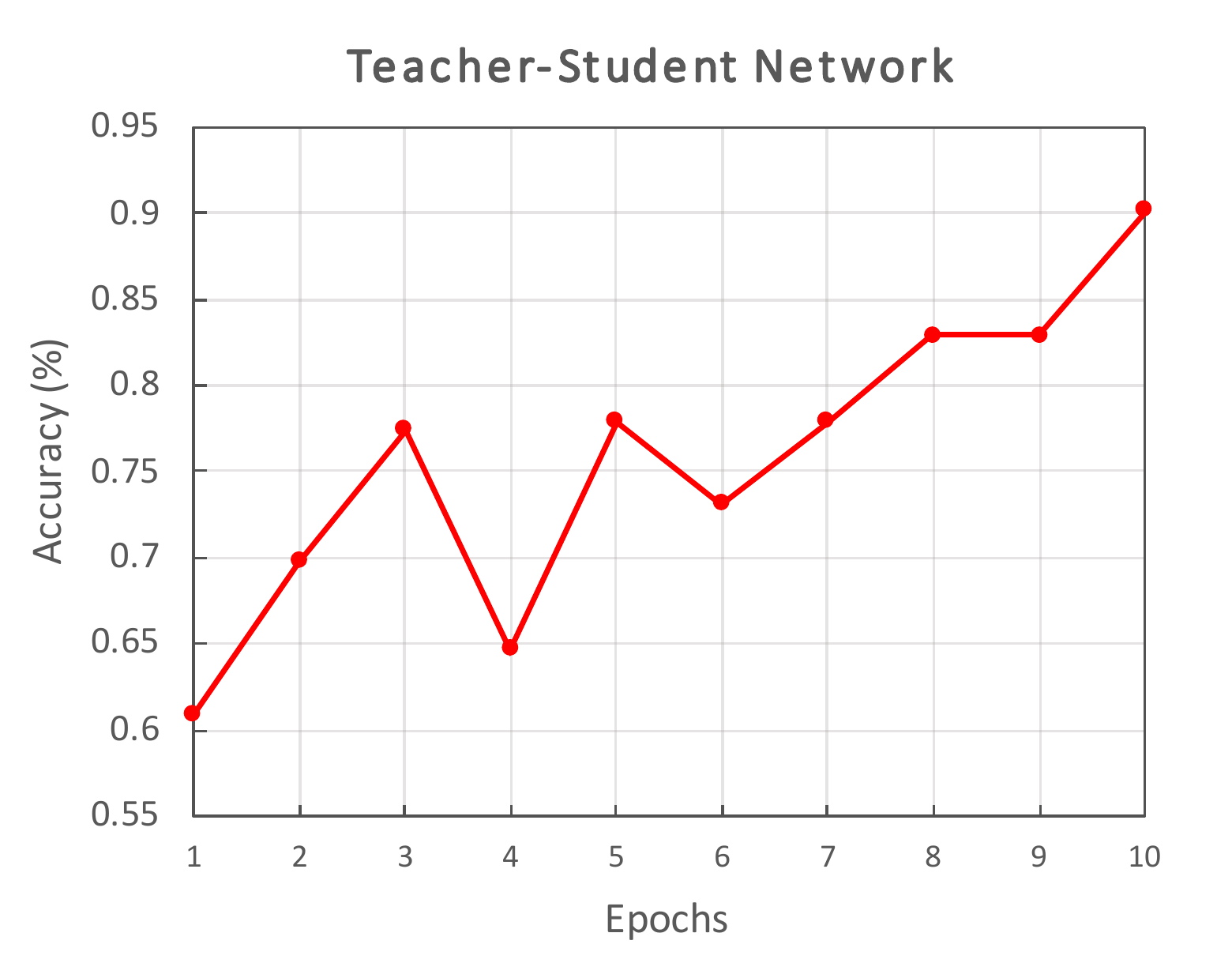}       
    \label{fig:tsbasic}
}
\hspace{5pt}
\subfigure[The comparisons with the predictions from our framework]{
    \includegraphics[width=0.31\linewidth, keepaspectratio]{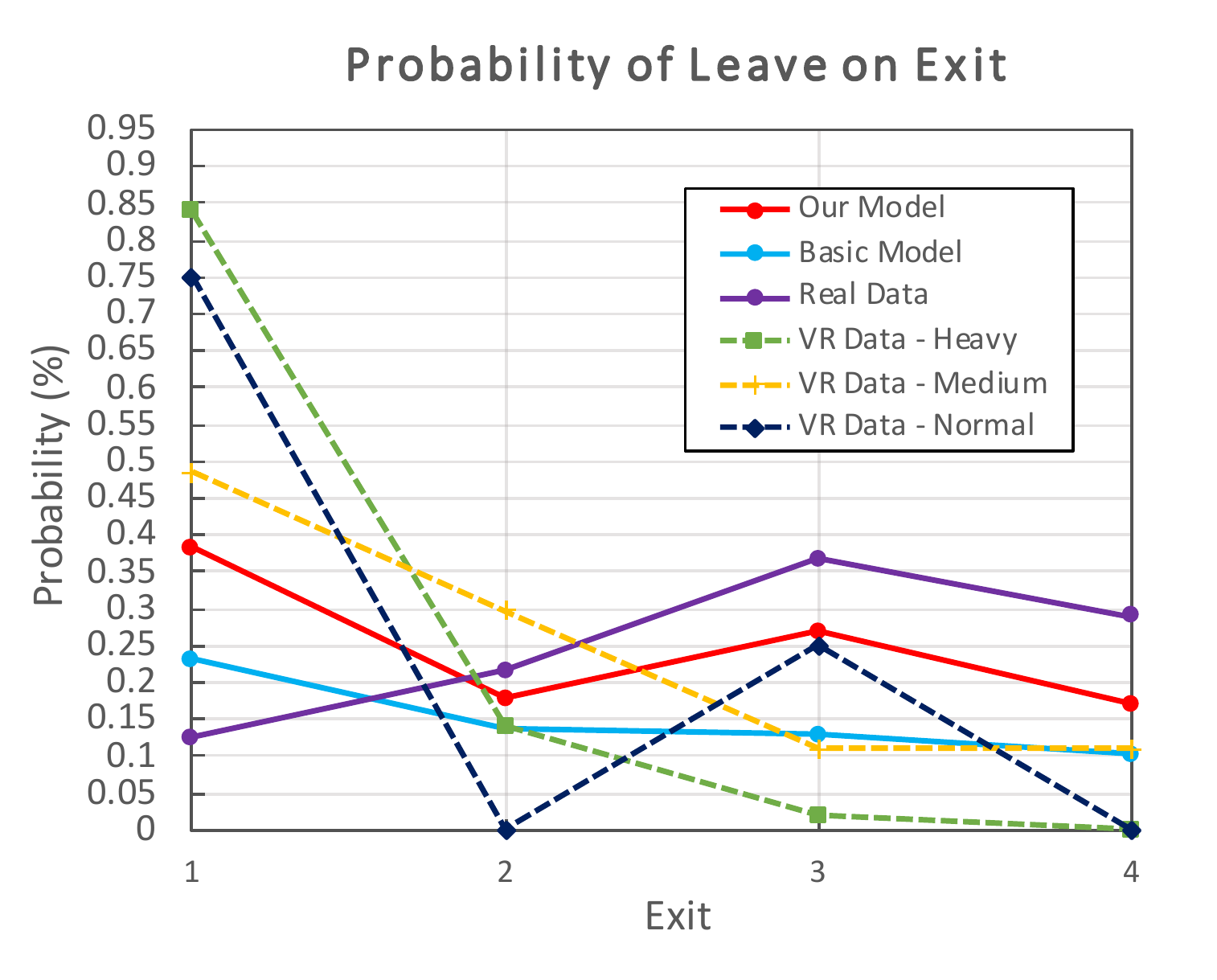}       
    \label{fig:comp_result}
}
\caption[Optional caption for list of figures]{The  prediction accuracy of our framework. Top two: shows the prediction accuracy of teacher and student networks on testing set. Bottom: shows the prediction accuracy of our framework (teacher-student network) on the left;  on the right shows the probabilities of leave at exits calculated based on the predictions from our framework. We compared our results with the results from the Basic model and Real data, we also plotted our VR data for showing better insights.}
\label{fig:all_accu}
\end{figure*}

\textbf{Real Data} We calculated the real probabilities of  taking the exits from the data provided by the (name omitted to preserve anonymity). Given the traffic volumes captured at the four exits, the probability of taking  an  exit is computed by,
\begin{equation}
P_{e} = \dfrac{V_{e}}{\mathlarger{\sum}\limits_{i \in {E}} V_i}
\end{equation}
where $E$ is the set of exits, $V_e$ is the traffic volume at exit $e$.  The real probabilities for the four exits are shown in Fig.\ref{fig:comp_result}. The real probabilities  will be used for evaluating the accuracy of  the predictions by our framework.

\begin {table}[b!]
\caption {Classification accuracy (\%) of teacher network with various architectures on VR data}
\begin{center}
\begin{tabular}{ p{6.0cm}|M{1.0cm} }
 \hline
 Network Architecture & Accuracy\\
 \hline
 10n-0.25DP-30n-0.35DP-20n & 92.02 \\
 10n-0.25DP-50n-0.35DP-30n-0.15DP  & 93.65 \\
 10n-0.25DP-50n-0.35DP-50n-0.25DP-50n-0.15DP & 94.37 \\
 \textbf{10n-0.25DP-30n-0.35DP-20n-0.25DP-50n-0.45DP} & \textbf{94.95} \\
 \hline
\end{tabular}
\label{mlp_arc}
\end{center}
\end{table}

\subsection{Implementation details}
Table \ref{mlp_arc} shows the architectures of the different teacher networks that we considered in our experiments. During experimentation, we varied the number of the neurons in each layer activated by ReLU function as well as the number of Dropout and hidden layers. In each architecture,  the output from the last layer of the feature extraction network is  input into a 4-way softmax layer that transforms the logits to  a probability distribution  over the four exits.  
From Table \ref{mlp_arc}, we can see that the  teacher network architecture described in the last row has the best   prediction result of \textbf{94.95\%} for VR data. The notation like "10n-0.25DP" indicates that the network has a dense layer with 10 neurons followed by a Dropout layer with a ratio of 0.25. For student network, 
we implemented a small feature extraction network with two dense layers of 10 and 20 neurons, both layers activated by ReLU functions. No dropout layer was  used in the student network. Instead we added a Batch Normalization layer to its second dense layer, and a 4-way softmax layer on the top  of the last layer. The inputs for the teacher and student networks have both 12 dimensions.

In our experiments, we computed the softened logits output from the last layer of our feature extraction networks, and then the softened softmax output  is obtained by applying the softmax layer on the softened logits. The original ground truth data concatenated with the softened softmax outputs from teacher network  is used for training and testing the student network. The predictions from  the student network concatenated with its softened softmax outputs are used to compute the distillation loss in each iteration. Then  we use backpropagation  for updating the parameters of the student network using gradient descent.   Finally, the  standalone trained student network is used for inference.


\begin {table}[b!]
\caption {Comparison of classification accuracy (\%) of our framework with teacher and student networks on Basic data}
\begin{center}
\begin{tabular}{ p{2.0cm}|M{1.0cm} }
 \hline
 Network & Accuracy\\
 \hline
 Teacher & 97.93 \\
 Student  & 77.45 \\
 \textbf{Our Model} & \textbf{95.20} \\
 \hline
\end{tabular}
\label{final_accu}
\end{center}
\end{table}

\subsection{Evaluation}
We  first trained our teacher network on VR data. The  performance of the teacher network on the testing set of the VR data is shown  in Fig.\ref{fig:tvr} and in  Table~\ref{mlp_arc}. From Fig.\ref{fig:tvr}, we can see that the prediction accuracy of teacher network trained on the VR data  improves fast along with increased epochs before stabilizing after epoch $5$.  Table \ref{final_accu} shows that  the teacher network trained on  the VR data  achieves the best prediction accuracy of \textbf{97.93\%} on the test set of  the basic data.   For the  student network in Fig.\ref{fig:sbasic}, we can see  that its prediction accuracy on the test set of the basic data improves with increasing number of epochs. The accuracy peaks at epoch 4, decreases after that before achieving a positive slope again at epoch 5. The best accuracy achieved by the student is \textbf{77.45\%} (Table \ref{mlp_arc}). The early peaking is due to the smaller size and the simplicity of the student network.   Our framework uses knowledge distillation to direct the training of the student network on  the prior knowledge acquired by the teacher from the VR data. Fig.\ref{fig:tsbasic} shows that after the 6th epoch,   the prediction accuracy of the augmented model (teacher-student network) on the test set of the basic data  increases with a higher slope with  respect to number of epochs than the student network.  Initially, the prediction accuracy increases with increasing number of epochs as the ground truth loss (loss from hard target) dominates. Starting  at epoch 4, where we see a dip, the  loss from soft target starts dominating as the teacher guides the student. 

After knowledge distillation  from the teacher  network to the student network, the prediction accuracy of the student network on the test set of the basic data  abruptly improves:  it achieves a classification accuracy of \textbf{95.2\%} on the basic data (see Fig.\ref{fig:comp_result}). Based on the prediction  results from our framework on basic data, the probabilities an individual driver  taking  the exits is computed and plotted in  Fig.\ref{fig:comp_result}. It can be seen from Fig.\ref{fig:comp_result} the prediction accuracy of  our framework is better than the baseline model:  our results are closer to the real data and have similar trend  except at Exit 1.  The prediction  at Exit 1 was heavily dominated by a large number of  discrepancies in drivers actions as seen from the VR data. However,  overall, our framework shows better fidelity than the baseline route choice model. Thus, using knowledge distillation, we have augmented a baseline model with contextual information acquired from SCE to obtain a model with higher fidelity. 

\section{Conclusion}\label{con}
In  this paper, we proposed  a novel  approach for developing high-fidelity route choice models with increased predictive power by augmenting existing aggregate level models with contextual information obtained from SCE carried out in an IVE through the use of knowledge distillation. To this end, we presented a general end-to-end knowledge distillation framework that uses a multilayer perceptron  as a  feature extraction network
to provide a feature learning architecture for teacher and
student networks  and then transfers  knowledge from the former to the latter by
optimizing  distillation loss.  Experimental results have demonstrated that that the predictions of  the augmented models produced by our approach  are much closer to reality than that of  the baseline. 

\section*{Acknowledgment}
This research was supported by Transportation Consortium of South-Central States (Tran-SET) Award No 18ITSLSU09/69A3551747016. Any opinions, findings, and conclusions or recommendations expressed in this material are those of the author(s) and do not necessarily reflect the views of the sponsor. 

\bibliographystyle{IEEEtran}
\bibliography{main.bib}

\end{document}